\documentclass[10pt, a4paper]{article}
\usepackage{lrec}
\usepackage{multibib}
\newcites{languageresource}{Language Resources}
\usepackage{graphicx}
\usepackage{tabularx}
\usepackage{soul}
\usepackage{epstopdf}
\usepackage[latin1]{inputenc}

\usepackage{hyperref}
\usepackage{xstring}

\makeatletter
\renewcommand{\@seccntformat}[1]{\csname the#1\endcsname\quad}
\makeatother
\renewcommand{\thesection}{\arabic{section}}
\renewcommand{\thesubsection}{\thesection.\arabic{subsection}}

\usepackage{amsmath}
\usepackage{amsfonts}
\usepackage{listings} % for amr
\usepackage[dvipsnames]{xcolor}
\usepackage{bm} 
\usepackage{bclogo}
\usepackage[textwidth=1.5cm]{todonotes} % tooltip notes for communication
\usepackage{calc}	  % calculations over tikz coordinates	
\usetikzlibrary{calc} % calculations over tikz coordinates,
\usepackage{tikz}     % drawing nodes
\usetikzlibrary{arrows} % tikz arrows
\usepackage{pifont} % arrow in amr to clausal form translation
\usepackage{booktabs,multirow} % nice tables
\usepackage{array,ragged2e}    % newcolumntype
\usepackage[normalem]{ulem} % underline and stikeout
\usepackage{caption} 
\usepackage{relsize} % relative size of font
\usepackage{mymacros} % user's macros
\usepackage{multicol} % multicolumn structure

\newcommand{\drgvar}[1]{\textbf{#1}}
\newcommand{\strout}[1]{\sout{\mbox{#1}}}
\newcommand{\wsp}{\hspace{3mm}}
\newcommand{\posn}[2]{#1\kern-0.15em.\kern-0.15em#2}
\newcommand{\smatch}{\textsc{smatch}}
\newcommand{\spar}{\textsc{Spar}}
\newcommand{\amrdrs}{\textsc{amr2drs}}
\newcommand{\dmatch}{\mbox{\textsc{counter}}}
\newcommand{\matched}[1]{\textcolor{green!40!black}{#1}}
\newcommand{\nonmatched}[1]{\textcolor{red!70!black}{#1}}
\newcommand{\ntt}[2][1]{\textsmaller[#1]{\texttt{#2}}}
\definecolor{pmbBlue}{RGB}{14,163,172}

\title{
    Evaluating Scoped Meaning Representations}

\name{Rik van Noord, Lasha Abzianidze, Hessel Haagsma, Johan Bos}

\address{CLCG, University of Groningen\\
         \{r.i.k.van.noord, l.abzianidze, hessel.haagsma, johan.bos\}@rug.nl\\}

\abstract{
Semantic parsing offers many opportunities to improve natural language understanding. 
We present a semantically annotated parallel corpus for English, German, Italian, and Dutch where sentences are aligned with scoped meaning representations in order to capture the semantics of negation, modals, quantification, and presupposition triggers. The semantic formalism is based on Discourse Representation Theory, but concepts are represented by WordNet synsets and thematic roles by VerbNet relations. Translating scoped meaning representations to sets of clauses enables us to compare them for the purpose of semantic parser evaluation and checking translations. This is done by computing precision and recall on matching clauses, in a similar way as is done for Abstract Meaning Representations. 
We show that our matching tool for evaluating scoped meaning representations is both accurate and efficient. Applying this matching tool to three baseline semantic parsers yields F-scores between 43\% and 54\%. A pilot study is performed to automatically find changes in meaning by comparing meaning representations of translations. This comparison turns out to be an additional way of (i) finding annotation mistakes and (ii) finding instances where our semantic analysis needs to be improved.\\
\newline
\Keywords{parallel corpus, semantic annotation, discourse representation structure, evaluation, semantic scope} }

\begin{document}

\maketitleabstract

%██████████████ INTRO █████████████
\section{Introduction}

Semantic parsing is the task of assigning meaning representations to natural language expressions. Informally speaking, a meaning representation describes \emph{who did what to whom, when, and where, and to what extent this is the case or not}. 
The availability of open-domain, wide coverage semantic parsers has the potential to add new functionality, such as detecting contradictions, verifying translations, and getting more accurate search results. Current research on open-domain semantic parsing focuses on supervised learning methods, using large semantically annotated corpora as training data. 

However, there are not many annotated corpora available.
We present a parallel corpus annotated with formal meaning representations for English, Dutch, German, and Italian, and a way to evaluate the quality of machine-generated meaning representations by comparing them to gold standard annotations. 
Our work shows many similarities with recent annotation and parsing efforts around Abstract Meaning Representations, (AMR; Banarescu et al., 2013\nocite{amr:13}) in that we abstract away from syntax, use first-order meaning representations, and use an adapted version of \smatch{} \cite{smatch:13} for evaluation. However, we deviate from AMR on several points: meanings are represented by scoped meaning representations (arriving at a more linguistically motivated treatment of modals, negation, presupposition, and quantification), and the non-logical symbols that we use are grounded in WordNet (concepts) and VerbNet (thematic roles), rather than PropBank \cite{propbank:05}. We also provide a syntactic analysis in the annotated corpus, in order to derive the semantic analyses in a compositional way.

We make the following contributions:
\begin{itemize}\itemsep0mm
\item A meaning representation with explicit scopes that combines WordNet and VerbNet with elements of formal logic (Section~\ref{sec:smr}).
\item A gold standard annotated parallel corpus of formal meaning representations for four languages (Section~\ref{sec:PMB}).
\item A tool that compares two scoped meaning representations for the purpose of evaluation (Section~\ref{sec:dmatch} and Section~\ref{sec:usingdmatch}).
\end{itemize}

%████████████ SCOPED DRS FIGURE ████████████
\begin{figure*}
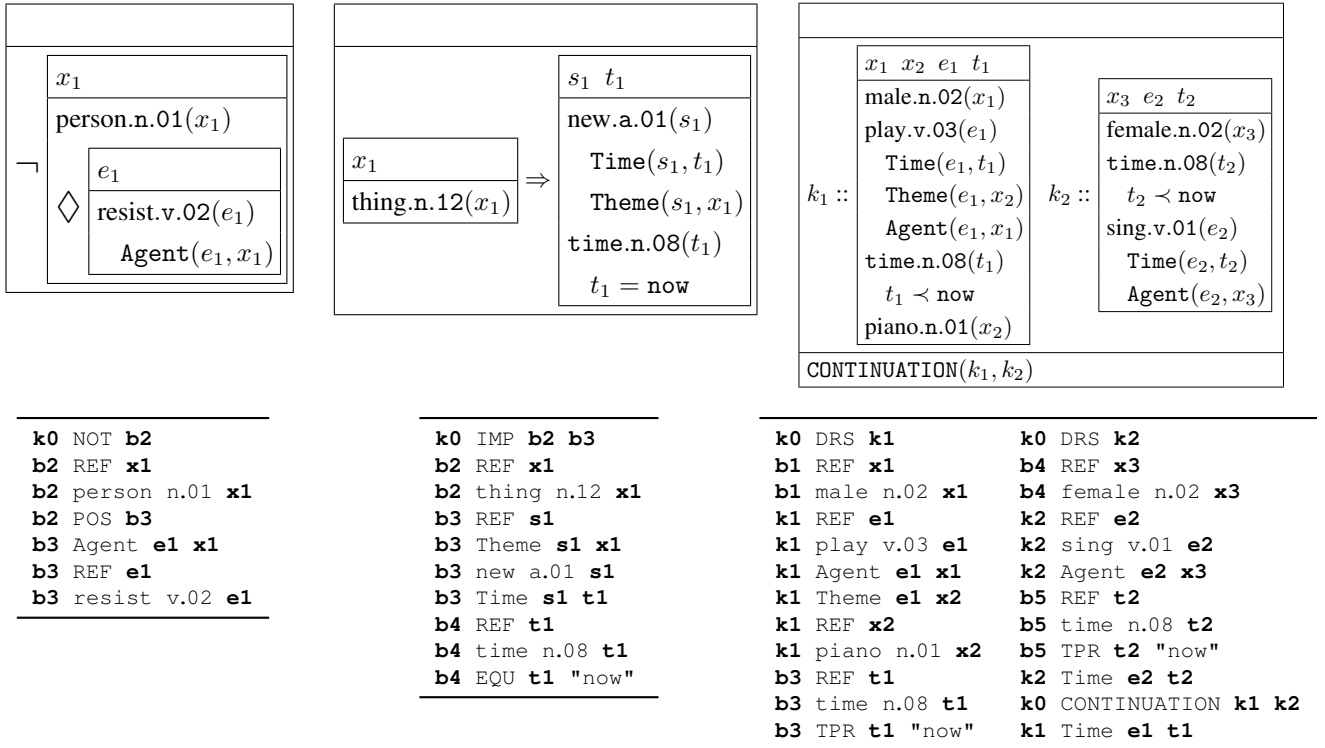

\hspace*{-3mm}
\begin{tabular}{c@{\kern3.5mm}c@{\kern-1.5mm}c}
%%%%%%%%%%%%%%%%%%%%%%
\textsmaller[1]{24/3221:~~\sym{No~one~can~resist.\phantom{y}}}
&
%\textsmaller[1]{00/2302:~~\sym{Everything~is~new.}}
\textsmaller[1]{00/2302:~~\sym{\grave{E}~tutto~nuovo.}}
&
\textsmaller[1]{00/3008:~~\sym{Hij~speelde~piano~en~zij~zong.}} 
\\[-1mm]
\scalebox{1}{
\renewcommand\arraystretch{1.3}
\drs[t]{}{\textlarger[2]{$\neg$}~\drs{$x_1$}{
    person$\sym{.n.01}$$(x_1)$\\
    \textlarger[2]{$\Diamond$}~\drs{$e_1$}{
        resist$\sym{.v.02}$$(e_1)$\\
        \wsp$\sym{Agent}(e_1, x_1)$    
    }}
}}
&
\scalebox{1}{
\renewcommand\arraystretch{1.3}
\drs[t]{}{\drs{$x_1$}{thing$\sym{.n.12}(x_1)$} 
        $\Rightarrow$
        \drs{$s_1$~~$t_1$}{
            new$\sym{.a.01}(s_1)$\\
            \wsp$\sym{Time}(s_1, t_1)$\\
            \wsp$\sym{Theme}(s_1, x_1)$\\
            $\sym{time.n.08}(t_1)$\\
            \wsp$t_1 = \sym{now}$
        }
}}
&
\scalebox{.9}{
\renewcommand\arraystretch{1.15}
\sdrs[t]{}{$k_1$ :: 
    \drs{$x_1$~~$x_2$~~$e_1$~~$t_1$}{
        male$\sym{.n.02}(x_1)$\\
        play$\sym{.v.03}(e_1)$\\
        \wsp$\sym{Time}(e_1, t_1)$\\
        \wsp$\sym{Theme}(e_1, x_2)$\\
        \wsp$\sym{Agent}(e_1, x_1)$\\
        $\sym{time.n.08}(t_1)$\\
        \wsp$t_1 \prec \sym{now}$\\
        piano$\sym{.n.01}(x_2)$
        } ~
    $k_2$ ::
    \drs{$x_3$~~$e_2$~~$t_2$}{
        female$\sym{.n.02}(x_3)$\\
        $\sym{time.n.08}(t_2)$\\
        \wsp$t_2 \prec \sym{now}$\\
        sing$\sym{.v.01}(e_2)$\\
        \wsp$\sym{Time}(e_2, t_2)$\\
        \wsp$\sym{Agent}(e_2, x_3)$
        } 
    }{$\sym{CONTINUATION}(k_1,k_2)$}
}
\\
\textsmaller[1]{\texttt{
\begin{tabular}[t]{l}\toprule
\drgvar{k0} NOT \drgvar{b2}\\
\drgvar{b2} REF \drgvar{x1}\\
\drgvar{b2} person \posn{n}{01} \drgvar{x1}\\
\drgvar{b2} POS \drgvar{b3}\\
\drgvar{b3} Agent \drgvar{e1} \drgvar{x1}\\
\drgvar{b3} REF \drgvar{e1}\\
\drgvar{b3} resist \posn{v}{02} \drgvar{e1}\\
\bottomrule
\end{tabular}
}}
&
\textsmaller[1]{\texttt{
\begin{tabular}[t]{l}\toprule
\drgvar{k0} IMP \drgvar{b2} \drgvar{b3}\\
\drgvar{b2} REF \drgvar{x1}\\
\drgvar{b2} thing \posn{n}{12} \drgvar{x1}\\
\drgvar{b3} REF \drgvar{s1}\\
\drgvar{b3} Theme \drgvar{s1} \drgvar{x1}\\
\drgvar{b3} new \posn{a}{01} \drgvar{s1}\\
\drgvar{b3} Time \drgvar{s1} \drgvar{t1}\\
\drgvar{b4} REF \drgvar{t1}\\
\drgvar{b4} time \posn{n}{08} \drgvar{t1}\\
\drgvar{b4} EQU \drgvar{t1} "now"\\
\bottomrule
\end{tabular}
}}
&
\textsmaller[1]{\texttt{
\begin{tabular}[t]{l@{\kern5mm}l}\toprule
\drgvar{k0} DRS \drgvar{k1} &
\drgvar{k0} DRS \drgvar{k2}\\
\drgvar{b1} REF \drgvar{x1} &
\drgvar{b4} REF \drgvar{x3}\\
\drgvar{b1} male \posn{n}{02} \drgvar{x1} &
\drgvar{b4} female \posn{n}{02} \drgvar{x3}\\
\drgvar{k1} REF \drgvar{e1} & 
\drgvar{k2} REF \drgvar{e2}\\
\drgvar{k1} play \posn{v}{03} \drgvar{e1} &
\drgvar{k2} sing \posn{v}{01} \drgvar{e2}\\
\drgvar{k1} Agent \drgvar{e1} \drgvar{x1} &
\drgvar{k2} Agent \drgvar{e2} \drgvar{x3}\\
\drgvar{k1} Theme \drgvar{e1} \drgvar{x2} &
\drgvar{b5} REF \drgvar{t2}\\
\drgvar{k1} REF \drgvar{x2} &
\drgvar{b5} time \posn{n}{08} \drgvar{t2}\\
\drgvar{k1} piano \posn{n}{01} \drgvar{x2} & 
\drgvar{b5} TPR \drgvar{t2} "now"\\
\drgvar{b3} REF \drgvar{t1} & 
\drgvar{k2} Time \drgvar{e2} \drgvar{t2}\\
\drgvar{b3} time \posn{n}{08} \drgvar{t1} &
\drgvar{k0} CONTINUATION \drgvar{k1} \drgvar{k2}\\
\drgvar{b3} TPR \drgvar{t1} "now" &
\drgvar{k1} Time \drgvar{e1} \drgvar{t1} \\
\bottomrule
\end{tabular}
}}
\end{tabular}
\caption{Examples of PMB documents with their scoped meaning representations and the corresponding clausal form. The first two structures are basic DRSs while the last one is a segmented DRS.}
\label{fig:drs-scopes}
\end{figure*}
%████████████ SCOPED DRS END FIGURE ████████████

%██████████████ SMR █████████████
\section{Scoped Meaning Representations}\label{sec:smr}

\subsection{Discourse Representation Structures}\label{ssec:drs}

%\noindent

The backbone of the meaning representations in our annotated corpus is formed by the Discourse Representation Structures (DRS) of Discourse Representation Theory \cite{kampreyle:drt}. Our version of DRS integrates WordNet senses \cite{wordnet}, adopts a neo-Davidsonian analysis of events employing VerbNet roles \cite{Bonial:11}, and includes an extensive set of comparison operators.
More formally, a DRS is an ordered pair of a set of variables (discourse referents) and a set of conditions. There are basic and complex conditions. Terms are either variables or constants, where the latter ones are used to account for indexicals \cite{indexicals:17}. 
Basic conditions are defined as follows:

\begin{itemize}\itemsep0mm
    \item If W is a symbol denoting a WordNet concept and x is a term,
          then W(x) is a basic condition;
    \item If V is a symbol denoting a thematic role 
          and x and y are terms, then V(x,y) is a basic condition;
    \item If x and y are terms,
          then x$=$y, 
               x$\neq$y,
               x$\sim$y, 
            x$<$y, 
             x$\leq$y, 
             x$\prec$y, and 
              x$\bowtie$y
               are basic conditions formed with comparison operators.
\end{itemize}

%\noindent

WordNet concepts are represented as word\sym{.POS.SenseNum}, denoting a unique synset within WordNet. Thematic roles, including the VerbNet roles, always have two arguments and start with an uppercase character. Complex conditions introduce scopes in the meaning representation. They are defined using logical operators as follows:

\begin{itemize}\itemsep0mm
    \item If B is a DRS, then $\lnot$B, $\Diamond$B, $\Box$B are complex conditions;
    \item If x is a variable, and B is a DRS, then x:B is a complex condition;
    \item If B and B' are DRSs, then B$\Rightarrow$B' and B$\lor$B' are complex conditions.
\end{itemize}

%\noindent
Besides basic DRSs, we also have segmented DRSs, following \newcite{asher:drt} and \newcite{asherlascarides}.
Hence, DRSs are formally defined as follows:

\begin{itemize}\itemsep0mm
    \item  If D is a (possibly empty) set of discourse referents, and C a (possibly empty) set of DRS-conditions, then $<$D,C$>$ is a (basic) DRS;
    \item If B is a (basic) DRS, and B' a DRS, then B$\downarrow$B' is a (segmented) DRS;
    \item If U is a set of labelled DRSs, and R a set of discourse relations, then $<$U,R$>$ is a (segmented) DRS.
\end{itemize}

DRSs can be visualized in different ways. While the compact linear format saves space, the box notation increases readability. 
In this paper we use the latter notation. 
The examples of DRSs in the box notation are presented in Figure\,\ref{fig:drs-scopes}.

%\medskip
%%\begin{equation}
%\scalebox{.89}{$
%    \Big[\,\Big|
%        \neg\big[x\big|
%            \text{person}\sym{.n.01}(x), ~ 
%            \Diamond[e_1|
%                \text{resist}\sym{.v.02}(e_1), 
%                \sym{Agent}(e_1, x)
%            ]
%        \big]
%    \Big]
%$}
%%\label{eq:one_line_drs}
%%\end{equation}
%\medskip

However, for evaluation and comparison purposes, we convert a DRS into a flat clausal form, i.e. a set of clauses.
This is carried out by using the labels for DRSs as introduced in  \newcite{venhuizen2015PhDthesis} and \newcite{venhuizen2018discourse}, and breaking down the recursive structure of DRS by assigning them a label of the DRS in which they appear.
Let t, t', and t'' be meta-variables ranging over DRSs or terms. Let $\cal{C}$ be a set of WordNet concepts, 
$\cal{T}$ a set of the thematic roles, and 
$\cal{O}$ the set of DRS operators (REF, NOT, POS, NEC, EQU, NEQ, APX, LES, LEQ, TPR, TAB, IMP, DIS, PRP, DRS).
%\addnote{missing subset\_of relation? JB: no!}
The resulting clauses are then of the form \mbox{t R t'} or \mbox{t R t' t''} where R $\in \cal{C}\cup\cal{T}\cup\cal{O}$.
The result of translating DRSs to sets of clauses is shown in Figure\,\ref{fig:drs-scopes}.
In a clausal form, it is assumed that different variables are represented with different variable names and vice versa.
Due to this, before translating a DRS to a clausal form, different discourse referents in the DRS must be represented with different variable names. 
This assumption significantly simplifies the matching process between clausal forms (Section \ref{sec:dmatch}) and makes it possible to recover the original box notation of a DRS from its clausal form.

%███████████████
\subsection{Comparing DRSs to AMRs}\label{ssec:drs_vs_amr}

Since DRSs in a clausal form come close to the triple notation of AMRs \cite{smatch:13}, and both aim to model meaning of natural language expressions, it is instructive to compare these two meaning representations.
The main difference between AMRs and DRSs is that the latter ones have explicit scopes (boxes) and scopal operators such as negation.
Due to the presence of scope in DRSs, their clauses are more complex than AMR triples. The length of DRS clauses varies from three to four, in contrast to the constant length of AMR triples. Additionally, DRS clauses contain two different types of variables, for scopes and discourse referents, whereas AMR triples have just one type.

Unlike AMRs, DRSs model tense. In general, the tense related information is encoded in a clausal form with three additional clauses, which express a WordNet concept, semantic role and a comparison operator. 
In order to give an intuition about the diversity of clauses in DRSs, Table~\ref{tab:dist} shows a distribution of various types of clauses in a corpus of DRSs (see Section\,\ref{sec:PMB}).
%Given that DRSs model tense, there is at least one comparison operator per declarative sentence that accounts for tense.
Since every logical operator carries a scope, their number represents a lower bound of the number of scopes in the meaning representations.
In addition to logical operators, scopes are introduced by presupposition triggers like proper names or pronouns.

\begin{table}[t]
\centering
\caption{\label{tab:dist}Distribution of clause types for 2,049 gold DRSs.}
\scalebox{.92}{
\begin{tabular}{@{\,}l@{~~}l@{~~}l@{~}r@{\,}}
\toprule
\textbf{Type} & \textbf{Description} & \textbf{Example} & \textbf{Total} %& \textbf{Unique} & \textbf{\%}
\\ \midrule
REF & Discourse referent
& \ntt{\drgvar{b3} REF \drgvar{x2}}  & 7,592     %&  & 
\\
NOT & Negation
& \ntt{\drgvar{b1} NOT \drgvar{b2}}  & 204      %&  & 
\\
POS & Possibility ($\Diamond$)              
& \ntt{\drgvar{b4} POS \drgvar{b5}}  & 55         %&  & 
\\
NEC & Necessity ($\Box$)   
& \ntt{\drgvar{b2} NEC \drgvar{b3}}  & 14         %&  & 
\\
IMP & Implication ($\Rightarrow$)
& \ntt{\drgvar{b1} IMP \drgvar{b2} \drgvar{b3}} &   104        %&  & 
\\
%DIS &\ntt{\drgvar{b2} DIS \drgvar{b3} \drgvar{b4}}  
%&0         %&  & 
%\\
PRP & Proposition ($:$)
& \ntt{\drgvar{b1} PRP \drgvar{x6}} & 50         %&  & 
\\
REL & Discourse relation
& \ntt{\drgvar{b1} CONTINUATION \drgvar{b2}} & 71         %&4 & 
\\
DRS & DRS as a condition 
& \ntt{\drgvar{b4} DRS \drgvar{b5}} & 84         %&  & 
\\
Compare & Comparison operators               
& \ntt{\drgvar{x1} APX \drgvar{x2}} &2,100 %&5 &
\\
Concept & WordNet senses
&\ntt{\drgvar{b2} hurt \posn{v}{02} \drgvar{e3}} &7,545%&1,767&
\\
Role & Semantic roles 
&\ntt{\drgvar{b2} Agent \drgvar{e3} \drgvar{x4}} &7,516   %&48&
\\
\bottomrule
\end{tabular}
}
%\caption{\label{tab:dist}Distribution of clause types for 2,049 gold DRSs.}
\end{table}

\begin{figure}[h]
  \centering
  \includegraphics[scale=0.445]{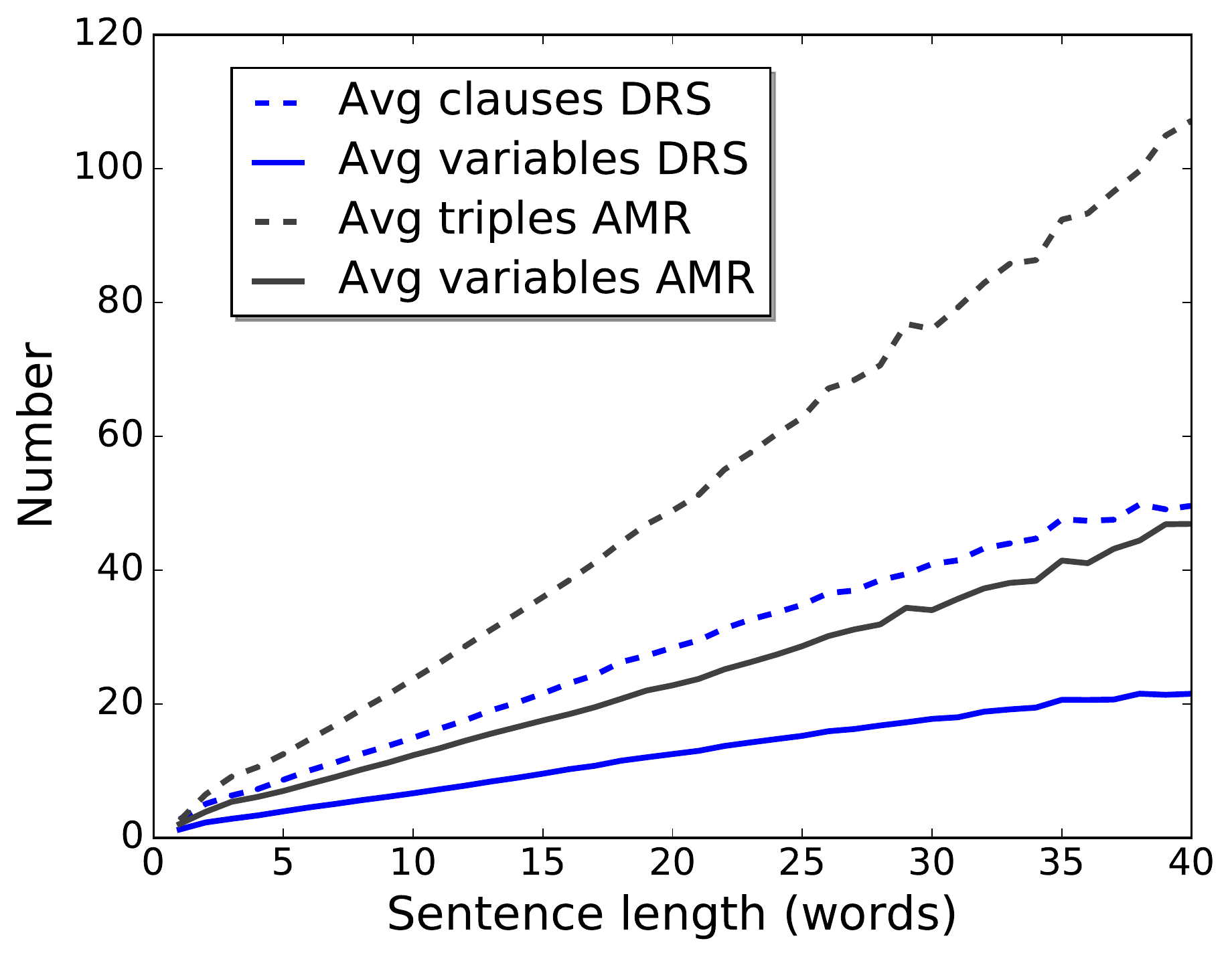}
  \caption{\label{fig:senlength}Comparison of the number of triples/clauses and variables between AMRs and DRSs for sentences of different length.}
\end{figure}

To make a meaningful comparison between AMRs and DRSs in terms of size, we compare the DRSs of 250,000 English sentences from the Parallel Meaning Bank (PMB; Abzianidze et al., 2017\nocite{PMBshort:2017}) to AMRs of the same sentences, produced by the state-of-the-art AMR parser from \newcite{clinAMR:17}.  
%used the state-of-the-art AMR parser from \newcite{clinAMR:17} to parse 250,000 English sentences from . Similar to the AMRs, DRSs for these sentences are produced by a semantic parser (Boxer) and not manually corrected.
Statistics of the comparison are shown in Figure\,\ref{fig:senlength}. On average, DRSs are about twice as large as AMRs, in terms of the number of clauses as well as the number of unique variables. 
This is obviously due to the explicit presence of scope in the meaning representation.
However, for both meaning representations the number of clauses and variables increase linearly with sentence length. %, as shown in Figure\,\ref{fig:senlength}.

\begin{figure*}
\centering
{\fboxsep=0mm%padding thickness
 \fboxrule=2pt%border thickness
 \fcolorbox{pmbBlue}{white}{\includegraphics[scale=0.32]{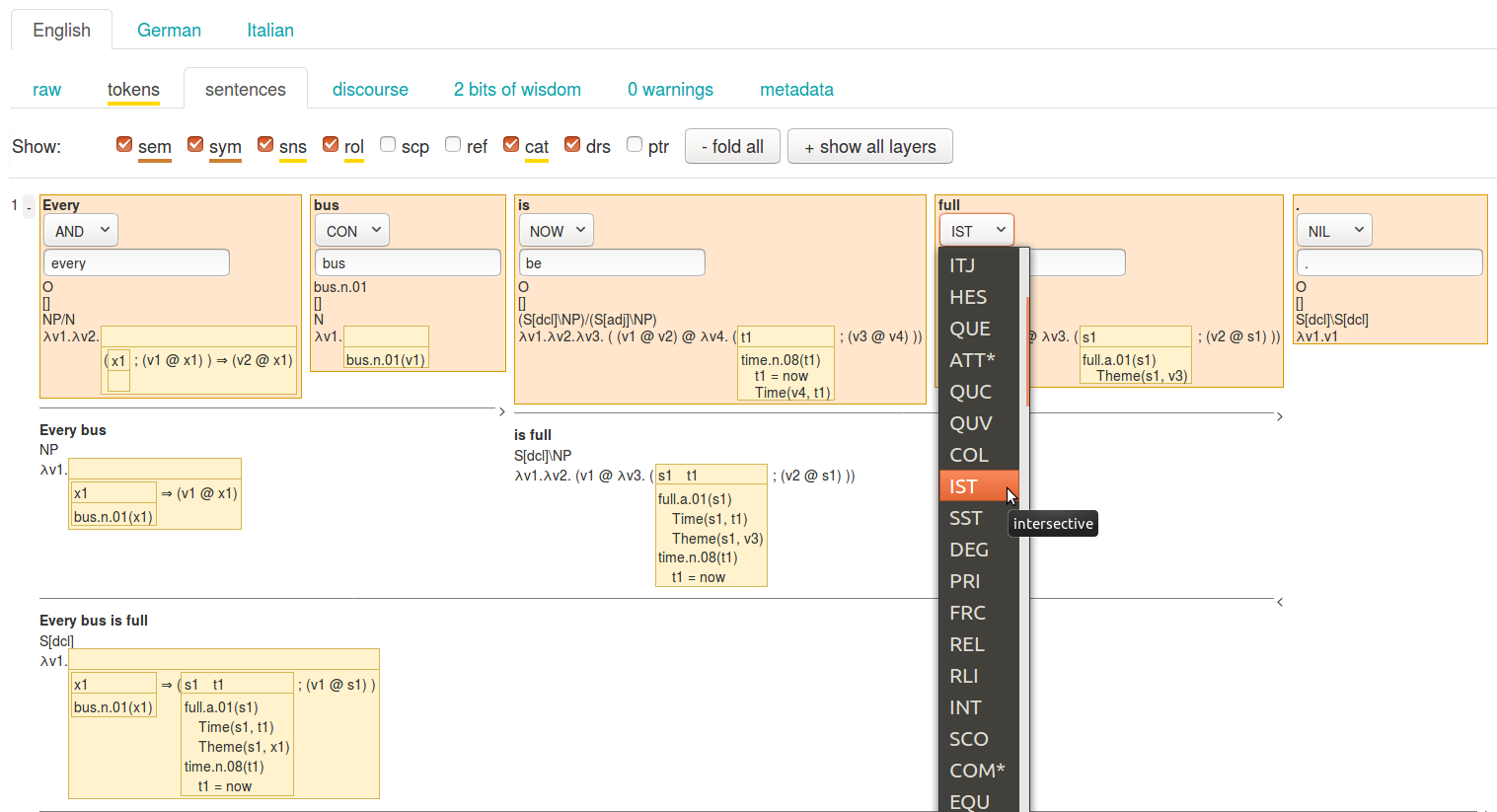}}
}
\caption{The edit mode of the PMB explorer: semantic tag ($\mathtt{sem}$) and symbol ($\mathtt{sym}$) layers of the document are bronze and therefore editable, while the word sense ($\mathtt{sns}$), semantic role ($\mathtt{rol}$) and CCG category ($\mathtt{cat}$) layers are gold and uneditable.}
\label{fig:explorer}
\end{figure*}
%██████████ Explorer screenshot ██████████

%██████████████ PMB █████████████
\section{The Parallel Meaning Bank}
\label{sec:PMB}

%\noindent

The scoped meaning representations, integrating word senses, thematic roles, and the list of operators, form the final product of our semantically annotated corpus: the Parallel Meaning Bank. The PMB is a semantically annotated corpus of English texts aligned with translations in Dutch, German and Italian \cite{PMBshort:2017}. 
It uses the same framework as the Groningen Meaning Bank \cite{Bos2017GMB}, but aims to abstract away from language-specific annotation models.
There are five annotation layers present in the PMB:
segmentation of words,
multi-word expressions and sentences \cite{elephant}, 
semantic tagging \cite{Bjervaetal:16,semantic-tagset:17}, 
syntactic analysis based on CCG \cite{lewisSteedman:14}, 
word senses based on WordNet \cite{wordnet}, and thematic role labelling \cite{BosEvangNissim2012}. 
The semantic analysis for English is projected on the other languages, to save manual annotation efforts \cite{evang-thesis,evang2016}. 
%Finally, the English meaning representations are projected on those for Dutch, Italian, German using the method described in \cite{evang2016}. %% This is not true anymore, right? We have language-specific parsing models for each language. JB: this is still true.
All the information provided by these layers is combined into a single meaning representation using the semantic parser Boxer \cite{boxer}, in the form of Discourse Representation Structures.
Note that the goal is to produce annotations that capture the most probable interpretation of a sentence; no ambiguities or under-specification techniques are employed.

\begin{table}[t]
\centering
\caption{Statistics of the first PMB release.\label{tab:release}}
\begin{tabular}{lrrr}
\toprule
                 & \textbf{Documents} & \textbf{Sentences} & \textbf{Tokens}  \\ \midrule
\textbf{English} & 2,049              & 2,057              & 11,664           \\
\textbf{German}  & 641                & 642                & 3,430           \\
\textbf{Italian} & 387                & 387                & 1,944            \\
\textbf{Dutch}   & 394                & 395                & 2,268           \\ \bottomrule
\end{tabular}
\end{table}

At each step in this pipeline, a single component produces the automatic annotation for all four languages, using language-specific models. 
Human annotators can correct machine output by adding `Bits of Wisdom' \cite{gmb:eacl}. These corrections serve as data for training better models, and create a gold standard annotated subset of the data. Annotation quality is defined per layer and language, at three levels: bronze (fully automatic), silver (automatic with some manual corrections), and gold (fully manually checked and corrected).  
If all layers are marked as gold, it follows that the resulting DRS can be considered gold standard, too.

%%% Gold standard data
%\section{Gold-standard Data} %Manually annotated data?
%\label{sec:gold}

The first public release\footnote{\url{http://pmb.let.rug.nl/data.php}} of the PMB contains gold standard scoped meaning representations for over 3,000 sentences in total (see Table~\ref{tab:release}).
%\addnote{LA: Table captions need to be in the bottom according to LREC the stylesheet.}
The release includes mainly relatively short sentences involving several semantic scope phenomena. 
A detailed distribution of clause types in the dataset is given in Table\,\ref{tab:dist}.    
%We choose to release the corpus to the community in a controlled way, given the difficulty of the semantic parsing task.
A larger amount of texts and more complex linguistic phenomena will be included in future releases. 

In addition to the released data, the PMB documents are publicly accessible through a web interface, called the PMB explorer.%
\footnote{\url{http://pmb.let.rug.nl/explorer}}
In the explorer, visitors can view natural language texts with several layers of annotations and compositionally derived meaning representations, and, after registration, edit the annotations.
It is also possible to use a word or a phrase search to find certain words or constructions with their semantic analyses.
%In order to edit annotations of a PMB document, a visitor only needs to register at the web site.
Figure~\ref{fig:explorer} shows the PMB explorer with the semantic analysis of a sentence in the edit mode.

%\addnote{Rik: if this won't fit maybe remove the unique column? Is not so interesting anyway}

%\addnote{Rik: This table was too large. Fixed it by changing unique to uniq but is that OK?}

% Old piece from application section
% In the gold DRGs, every noun, verb and adjective is annotated with its correct sense. Similar to \texttt{SMATCH} we evaluate on whole entities, meaning that if only the sense is incorrect, the whole concept is treated as incorrect. Table \ref{tab:results_sense_boxer} shows the impact of word-sense disambiguation on the performance. The first result is a baseline system and also what Boxer does currently, simply assigning the first sense to each concept.\footnote{In the set of gold DRGs, for the 3,395 entities that have a sense, 2,550 had the first sense.} This can be compared to ignoring the sense completely (thus always choosing the correct sense) and never outputting the correct sense. The latter enables us to determine the impact of concept identification, since all produced concept will be incorrect. We see that there might be not much to gain by improving word sense disambiguation: the baseline system is already close to perfect sense performance. 

%██████████████ Evaluation █████████████
\section{Matching Scoped Representations}
\label{sec:dmatch}

\subsection{Evaluation by Matching}
\label{ssec:eval-by-match}

In the context of the Parallel Meaning Bank there are two main reasons to verify whether two scoped meaning representations capture the same meaning or not:
(1) to be able to evaluate semantic parsers that produce scoped meaning representations by comparing gold-standard DRSs to system output; and (2) to check whether translations are meaning-preserving; a discrepancy in meaning between source and target could indicate a mistranslation.

The ideal way to compare two meaning representations would be one based on inference. This can be implemented by translating DRSs to first-order formulas and using an off-the-shelf theorem prover to find out whether the two meanings are logically equivalent \cite{blackburnbos:2005}. This method can compare meaning representation that have different syntactic structures but still are equivalent in meaning.  The disadvantage of this approach is that it yields just a binary answer: if a proof is found the meanings are the same, else they are not.

%Explain advantages and disadvantages - explain why variable matching is necessary (use example?)

\begin{figure}[t]
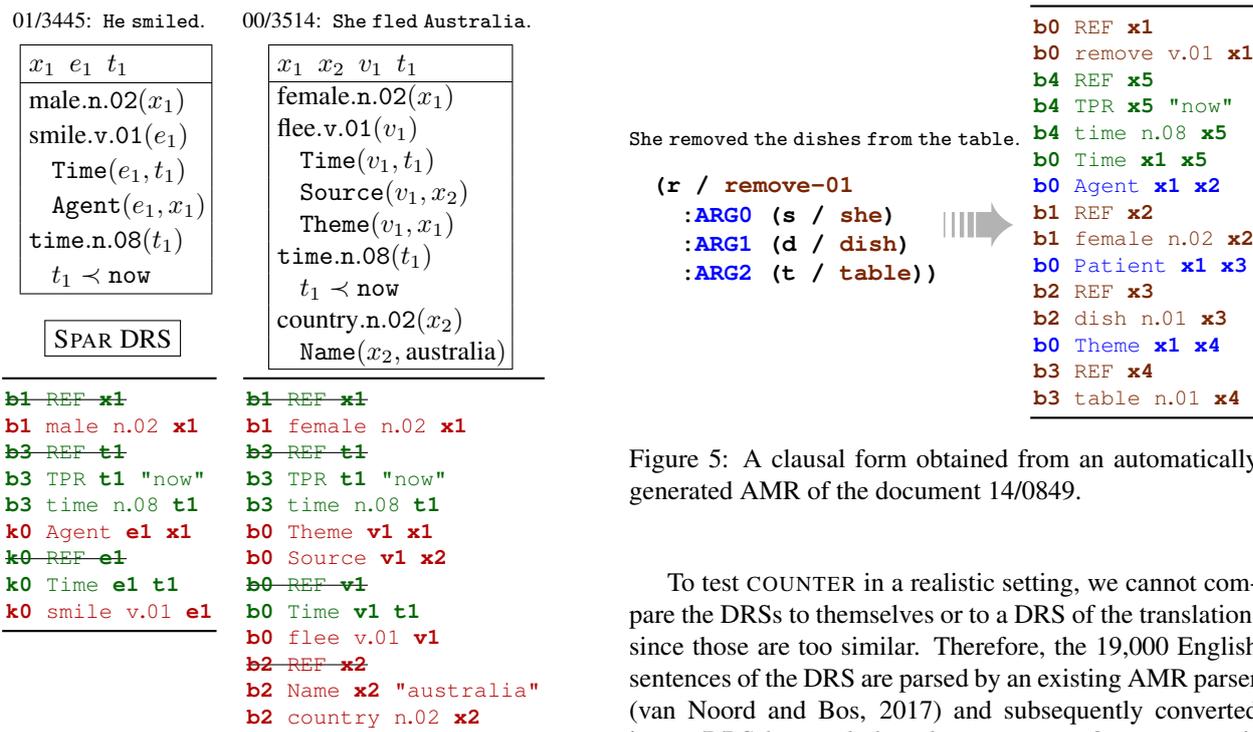

\centering
\begin{tabular}{@{}c@{}@{}c@{}}
%%%%%%%%%%%%%%%%%%%%%%
\textsmaller[1]{01/3445:~~\sym{He~smiled.}}
&
\textsmaller[1]{00/3514:~~\sym{She~fled~Australia.}}
\\[-3mm]
\scalebox{1}{
\begin{tabular}[t]{@{}c@{}}
\renewcommand\arraystretch{1.1}
\drs[t]{$x_1$ ~$e_1$ ~$t_1$}{
        male$\sym{.n.02}(x_1)$\\
        smile$\sym{.v.01}(e_1)$\\
        \wsp$\sym{Time}(e_1, t_1)$\\
        \wsp$\sym{Agent}(e_1, x_1)$\\
        $\sym{time.n.08}(t_1)$\\
        \wsp$t_1 \prec \sym{now}$
}\\
\raisebox{-3mm}{\fbox{\spar{} DRS}}
\end{tabular}} 
&
\scalebox{1}{
\drs[t]{$x_1$ ~$x_2$ ~$v_1$ ~$t_1$}{
        female$\sym{.n.02}(x_1)$\\
        flee$\sym{.v.01}(v_1)$\\
        \wsp$\sym{Time}(v_1, t_1)$\\
        \wsp$\sym{Source}(v_1, x_2)$\\
        \wsp$\sym{Theme}(v_1, x_1)$\\
        $\sym{time.n.08}(t_1)$\\
        \wsp$t_1 \prec \sym{now}$\\
        country$\sym{.n.02}(x_2)$\\
        \wsp$\sym{Name}(x_2,\text{australia})$
}} 
\\[-3mm]
\textsmaller[1]{\texttt{
\begin{tabular}[t]{@{\,}l@{\,}}\toprule
\strout{\matched{\drgvar{b1} REF \drgvar{x1}}}\\
\nonmatched{\drgvar{b1} male \posn{n}{02} \drgvar{x1}}\\
\strout{\matched{\drgvar{b3} REF \drgvar{t1}}}\\
\matched{\drgvar{b3} TPR \drgvar{t1} "now"}\\
\matched{\drgvar{b3} time \posn{n}{08} \drgvar{t1}}\\
\nonmatched{\drgvar{k0} Agent \drgvar{e1} \drgvar{x1}}\\
\strout{\matched{\drgvar{k0} REF \drgvar{e1}}}\\
\matched{\drgvar{k0} Time \drgvar{e1} \drgvar{t1}}\\
\nonmatched{\drgvar{k0} smile \posn{v}{01} \drgvar{e1}}\\
\bottomrule
\end{tabular}
}}
&
\textsmaller[1]{\texttt{
\begin{tabular}[t]{@{\,}l@{\,}}\toprule
\strout{\matched{\drgvar{b1} REF \drgvar{x1}}}\\
\nonmatched{\drgvar{b1} female \posn{n}{02} \drgvar{x1}}\\
\strout{\matched{\drgvar{b3} REF \drgvar{t1}}}\\
\matched{\drgvar{b3} TPR \drgvar{t1} "now"}\\
\matched{\drgvar{b3} time \posn{n}{08} \drgvar{t1}}\\
\nonmatched{\drgvar{b0} Theme \drgvar{v1} \drgvar{x1}}\\
\nonmatched{\drgvar{b0} Source \drgvar{v1} \drgvar{x2}}\\
\strout{\matched{\drgvar{b0} REF \drgvar{v1}}}\\
\matched{\drgvar{b0} Time \drgvar{v1} \drgvar{t1}}\\
\nonmatched{\drgvar{b0} flee \posn{v}{01} \drgvar{v1}}\\
\strout{\nonmatched{\drgvar{b2} REF \drgvar{x2}}}\\
\nonmatched{\drgvar{b2} Name \drgvar{x2} "australia"}\\
\nonmatched{\drgvar{b2} country \posn{n}{02} \drgvar{x2}}\\
\bottomrule
\end{tabular}
}}\\
\end{tabular}

\caption{The \spar{} DRS (Section\,\ref{ssec:sempar}) matches the DRS of 00/3514 PMB document with an F-score of 54.5\%. If redundant REF-clauses are ignored, the F-score drops to 40\%. 
These results are achieved with the help of the mapping \{\ntt[0]{k0}$\mapsto$\ntt[0]{b0}, \ntt[0]{e1}$\mapsto$\ntt[0]{v1}\}.
}
    \label{fig:drs-compare}
\end{figure}

An alternative way of comparing meaning representations is comparing the corresponding clausal forms by computing precision and recall over matched clauses \cite{allen:step2008}. The advantage of this approach is that it returns a score between 0 and 1, preferring meaning representations that better approximate the gold standard over those that are completely different. Since the variables of different clausal forms are independent from each other, the comparison of two clausal forms boils down to finding a (partial) one-to-one variable mapping that maximizes intersection of the clausal forms.
For example, the maximal matching for the clausal forms in Figure \ref{fig:drs-compare} is achieved by the following partial mapping from the variables of the left form into the variables of the right one:
\{\ntt[0]{k0}$\mapsto$\ntt[0]{b0}, \ntt[0]{e1}$\mapsto$\ntt[0]{v1}\}.

For AMRs, finding a maximal matching is done using a hill-climbing algorithm called \smatch{} \cite{smatch:13}. This algorithm is based on a simple principle: it checks if a single change in the current mapping results in a better matching mapping. If this is the case, it continues with the new mapping. Otherwise, the algorithm stops and has arrived at the final mapping. This means that it can easily get stuck in local optima. To avoid this, \smatch{} does a predefined number of restarts of this process, where each restart starts with a new and random initial mapping. The first restart always uses a `smart' initial mapping, based on matching concepts.

%Explain \texttt{D-Match} and why we run into problems

%██████ DRS COMPARISON FIGURE

%██████ DRS COMPARISON END FIGURE

Our evaluation system, called \dmatch{}\footnote{\url{http://github.com/RikVN/DRS_parsing/}}, is a modified version of \smatch{}. Even though clausal forms do not form a graph and clauses consist of either three or four components, the principle behind the variable matching is the same. The actual implementation differs, mainly because \smatch{} was not designed to handle clauses with three variables, e.g. $\langle$\ntt[0]{k0 Agent e1 x1}$\rangle$. 
%An example of comparing DRSs and their F-score is given in Figure\,\ref{fig:drs-compare}.

In contrast to \smatch{}, \dmatch{} takes a set of clauses directly as input. \dmatch{} also uses two smart initial mappings, based on either role-clauses, like $\langle$\ntt[0]{k0 Agent e1 x1}$\rangle$, or concept-clauses, like $\langle$\ntt[0]{k0 smile v.01 e1}$\rangle$. 

Also specific to this method is the treatment of REF-clauses in the matching process. Before matching two DRSs, redundant REF-clauses are removed. A REF-clause $\langle$\ntt[0]{b1 REF x1}$\rangle$ is redundant if its discourse referent \ntt[0]{x1} occurs in some basic condition of the same DRS \ntt[0]{b1}. Figure~\ref{fig:drs-compare} shows some examples of redundant REF-clauses. Not removing these redundant clauses would lead to inflated matching scores since for each matched variable the corresponding REF-clause will also match. Comparison of the clausal forms in Figure\,\ref{fig:drs-compare} demonstrates this fact. Note that not all REF-clauses are redundant: if a discourse referent is declared outside the scope of negation or an other scope operator, the REF-clause is kept. This is very infrequent in our data, since only a single REF-clause was preserved in 2,049 examples.

%If we look at a few very large DRGs, it becomes clear that \texttt{SMATCH} was never designed to handle this. A single restart for DRGs with more than 1000 triples can already take more than an hour. In comparison, if we compare all 36,521 AMRs to themselves, with four restarts, the algorithm is finished in 33 minutes.

%Last paragraph of this section?

%Ideally, combine the two approaches (in theory): try semantic match. If proof, return 1. Else try syntactic match and return result. What we do in practice is simply the syntactic match. Or something in between (using hypernyms??) \addnote{Rik: this needs to be finished}

\subsection{Evaluating Matching}
\label{ssec:eval-match}

%\noindent

As we showed in Figure\,\ref{fig:senlength}, DRSs are about twice as large as AMRs. This increase in size might be problematic, since it increases the average runtime for comparing DRSs. Moreover, if there are more variables, more restarts might be needed to ensure a reliable score, again increasing runtime.

Therefore, our goal is that \dmatch{} gets close to optimal performance in reasonable time. Since we want to be sure that this also holds for longer sentences, we use a balanced data set. We take 1,000 DRSs produced by the semantic parser Boxer for each sentence length from 2 to 20 (punctuation excluded), resulting in a set of 19,000 DRSs.

\begin{figure}[t]
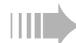

%\hspace*{-3mm}
\begin{tabular}{@{}c@{}c@{}}
%%%%%%%%%%%%%%%%%%%%%% 14/0849
\begin{minipage}[c]{5.1cm}
\textsmaller[1]{\sym{She~removed~the~dishes~from~the~table.}}\\[2mm]
\begin{tabular}{c@{}c@{}}
\lstset{
moredelim=[is][\bfseries\textcolor{blue}]{/*}{*/},
moredelim=[is][\bfseries]{<}{>},
moredelim=[is][\textcolor{Brown}]{"}{"}
}
\begin{lstlisting}[basicstyle={\ttfamily\footnotesize\bfseries}]
(<r> / "remove-01" 
  :/*ARG0*/ (<s> / "she") 
  :/*ARG1*/ (<d> / "dish") 
  :/*ARG2*/ (<t> / "table"))
\end{lstlisting}
&
\raisebox{-3mm}{\resizebox{10mm}{7mm}{\textcolor{black!30}{\ding{224}}}}   
%\raisebox{6.3cm}{\resizebox{2.5cm}{3cm}{\ding{223}}}
%\raisebox{6.3cm}{\resizebox{2.5cm}{3cm}{\ding{230}}}
\end{tabular}      
\end{minipage}
&
\textsmaller[1]{\texttt{
\begin{tabular}[c]{@{\,}l@{\,}}\toprule
\textcolor{Brown}{\drgvar{b0} REF \drgvar{x1}}\\
\textcolor{Brown}{\drgvar{b0} remove \posn{v}{01} \drgvar{x1}}\\
\matched{\drgvar{b4} REF \drgvar{x5}}\\
\matched{\drgvar{b4} TPR \drgvar{x5} "now"}\\
\matched{\drgvar{b4} time \posn{n}{08} \drgvar{x5}}\\
\matched{\drgvar{b0} Time \drgvar{x1} \drgvar{x5}}\\
\textcolor{blue}{\drgvar{b0} Agent \drgvar{x1} \drgvar{x2}}\\
\textcolor{Brown}{\drgvar{b1} REF \drgvar{x2}}\\
\textcolor{Brown}{\drgvar{b1} female \posn{n}{02} \drgvar{x2}}\\
\textcolor{blue}{\drgvar{b0} Patient \drgvar{x1} \drgvar{x3}}\\
\textcolor{Brown}{\drgvar{b2} REF \drgvar{x3}}\\
\textcolor{Brown}{\drgvar{b2} dish \posn{n}{01} \drgvar{x3}}\\
\textcolor{blue}{\drgvar{b0} Theme \drgvar{x1} \drgvar{x4}}\\
\textcolor{Brown}{\drgvar{b3} REF \drgvar{x4}}\\
\textcolor{Brown}{\drgvar{b3} table \posn{n}{01} \drgvar{x4}}\\
\bottomrule
\end{tabular}
}}%
% \raisebox{1mm}{
% \scalebox{.8}{
% \renewcommand\arraystretch{1.19}
% \drs[t]{$x_1$ ~$x_2$ ~$e_1$ ~$x_3$ ~$t_1$ \hspace{1mm}}{ 
% 		female$\sym{.n.02}(x_1)$\\
% 		remove$\sym{.v.01}(e_1)$\\
%         \wsp$\sym{Time}(e_1, t_1)$\\
%         \wsp$\sym{Source}(e_1, x_3)$\\
%         \wsp$\sym{Theme}(e_1, x_2)$\\
%         \wsp$\sym{Agent}(e_1, x_1)$\\
%         $\sym{time.n.08}(t_1)$\\
%         \wsp$t_1 \prec \sym{now}$\\
%         dish$\sym{.n.01}(x_2)$\\
%         table$\sym{.n.03}(x_3)$
% }}
% }
\end{tabular}
\caption{A clausal form obtained from an automatically generated AMR of the document 14/0849.}
\label{fig:amr2drs}
\end{figure}

To test \dmatch{} in a realistic setting, we cannot compare the DRSs to themselves or to a DRS of the translation, since those are too similar.
Therefore, the 19,000 English sentences of the DRS are parsed by an existing AMR parser \cite{clinAMR:17} and subsequently converted into a DRS by a rule-based system, \amrdrs{}, as motivated by \newcite{bos:16}.
An example of translating an AMR to a clausal form of a DRS is shown in Figure\,\ref{fig:amr2drs}. We convert AMR relations to DRS roles by employing a manually created translation dictionary, including rules for semantic roles (e.g. \ntt[0]{:ARG0} $\mapsto$ \ntt[0]{Agent} and \ntt[0]{:ARG1} $\mapsto$ \ntt[0]{Patient}) and pronouns (e.g. \ntt[0]{she} $\mapsto$ \sym{female.n.02}). Since AMRs do not contain tense information, past tense clauses%
\footnote{Past tense was chosen because it is the most frequent tense in the data set.}
are produced for the first verb in the AMR (see four tense related clauses in Figure\,\ref{fig:amr2drs}).
Also, since AMRs do not use WordNet synsets, all concepts  get a default first sense, except for concepts that are added by concept-specific rules, such as \sym{female.n.02} and \sym{time.n.08}.
%Also, only verbs have a sense in the AMR format, meaning that we treat all other concepts as nouns with a default first sense.\addnote{The verbs have coarse-grained senses that are different from WN senses.} 

%We do this by comparing the set of gold DRSs in the release to the set of DRSs produced by the semantic parser Boxer. Boxer utilizes the tagging layers described in Section \ref{sec:PMB}, but does not have access to the manual annotations that correct the output. Without these manual corrections, Boxer was not able to produce a valid DRS for 9 sentences, which were excluded from the experiment. 

%████████████ TRANSLATIONS END FIGURE ████████████

\begin{figure}[t]
  \centering
  \includegraphics[scale=0.445]{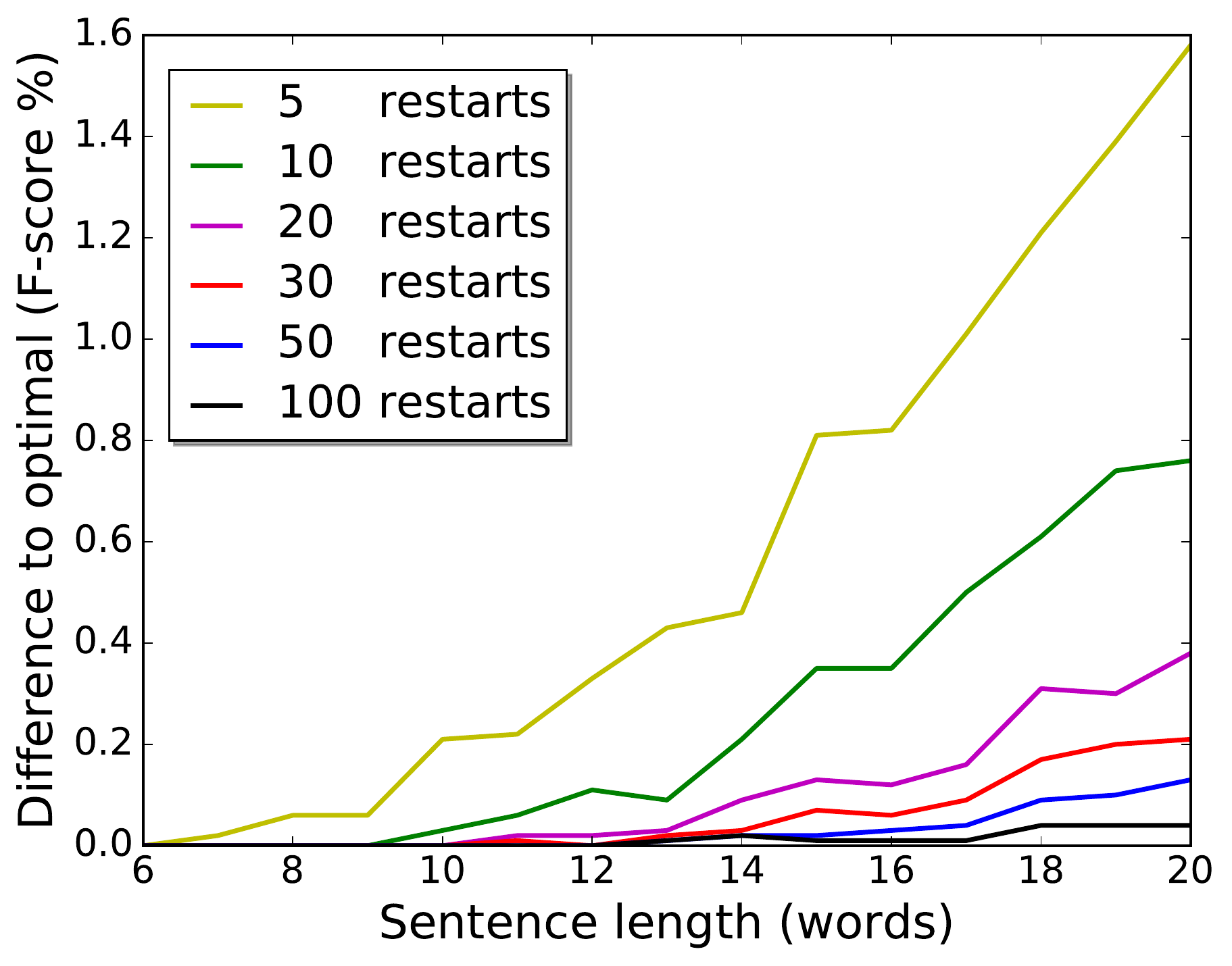}
  \caption{\label{fig:len_comparison}Comparison of the differences to the optimal \mbox{F-score} per sentence length for different number of restarts.}
\end{figure}

We compare the sets of DRSs using different numbers of restarts to find the best trade-off between speed and accuracy. 
%Note that in the cases were there is more than two restarts, the first two restarts always use the smart role and concept mapping.
The results are shown in Table \ref{tab:results}. %\footnote{All reported F-scores are averages of five runs of the \dmatch{} system. Note that we use a default version of Boxer here.}
The optimal scores are obtained using a Prolog script that performs an exhaustive search for the optimal mapping. 
As expected, increasing the number of restarts benefits performance. \newcite{smatch:13} consider four restarts the optimal trade-off between accuracy and speed, showing no improvement in F-score when using more than ten restarts.%
\footnote{However, we found that, in practice, \smatch{} still improves when using more restarts. Parsing the development set of the AMR dataset LDC2016E25 with the baseline parser of \newcite{clinAMR:17} yields an F-score of 55.0 for 10 restarts, but 55.4 for 100 restarts.} 
Contrary to \smatch{}, performance for \dmatch{} still increases with more than 4 restarts. In our case, it is a bit harder to select an optimal number of restarts, since this number depends on the length of the sentence, as shown in Figure~\ref{fig:len_comparison}. We see that for long sentences, 5 and 10 restarts are not sufficient to get close to the optimal, while for short sentences 5 restarts might be considered enough. In general, the best trade-off between speed and accuracy is approximately 20 restarts.

%We have to note, though, that these experiments were performed on a set of DRSs with simple, small sentences. DRSs for larger sentences will contain more clauses and variables (see Figure\,\ref{fig:senlength}) and will most likely need more restarts to reach optimal performance.

\begin{table}[ht!]
\centering
\caption{\label{tab:results}Results of comparing 19,000 Boxer-produced DRSs to DRSs produced by \amrdrs{}, for different number of restarts. For three or more restarts, we always use the smart role and concept mapping.}
\scalebox{.9}{
\begin{tabular}{r|ccc|r}
\toprule
\textbf{Restarts} & \textbf{P}\% & \textbf{R}\% & \textbf{F1}\% & \textbf{Time (h:m:s)}  \\ \midrule
(random) 1               & 27.20    & 22.71  & 24.75   &  4:19 
\\
(smart concepts) 1      & 27.45   & 22.92  & 24.98   & 4:35
\\
(smart roles) 1    & 27.27    & 22.76   & 24.81   & 4:37
\\
5                  & 30.25    & 25.25   & 27.53  & 19:33
\\
10               &  30.65  & 25.59  & 27.89   &  37:08
\\
20              & 30.84 & 25.75 & 28.07 &    1:10:13
\\
30              & 30.90   & 25.80  & 28.12   & 1:41:43
\\
50              & 30.94   & 25.83  & 28.16   &  2:41:38
\\
75              & 30.96   & 25.85  & 28.17   &  3:53:01
\\
100              & 30.97   & 25.85  & 28.18   & 5:01:25
\\\midrule
Optimal          & 30.98   & 25.86   & 28.19  & % excluding length 1
% SELECT COUNT(*), SUM(matched)/SUM(boxer_tuples), SUM(matched)/SUM(amr2drs_tuples), 2*SUM(matched)/(SUM(boxer_tuples)+SUM(amr2drs_tuples)) FROM `pmb_sen_promatch` WHERE type = 'noREF' AND sen_length != 1
\\\bottomrule                
\end{tabular}
}
\end{table}

%%%% Figure comparing different sentence lengths

% Ideas:

%\begin{itemize}

%\item baseline semantic parser SPAR.

%\end{itemize}

%\section{Putting \dmatch\ in action?}
\section{\dmatch\ in Action}
\label{sec:usingdmatch}
\subsection{Semantic Parsing}\label{ssec:sempar}

The first purpose of \dmatch{} is to evaluate semantic parsers for DRSs. Since this is a new task, there are no existing systems that are able to do this. Therefore, we show the results of three baseline systems \textsc{pmb\,pipeline}, 
%(with bare settings, i.e. without reference resolution, default WordNet senses and VerbNet roles),\addnote{Move to a footnote? Also Table 4 needs to be referred and said something about it} 
\spar{}, and \amrdrs{} (Subsection\,\ref{ssec:eval-match}).\footnote{\spar{} and \amrdrs{} are available at: \url{https://github.com/RikVN/DRS\_parsing/}}

The \textsc{pmb\,pipeline} produces a DRS via the pipeline of the tools used for automatic annotation of the PMB.%
\footnote{\url{http://pmb.let.rug.nl/software.php}}
This means that it has no access to manual corrections, and hence it uses the most frequent word senses and default VerbNet roles.
\spar{} is a trivial semantic `parser' which always outputs the DRS that is most similar to all other DRSs in the most recent PMB release (the left-hand DRS in Figure\,\ref{fig:drs-compare}). 

\begin{table}[t]
\centering
\caption{\label{tab:baselines}Comparison of three baseline DRS parsers to the gold-standard data set.}
\begin{tabular}{lrrr}
\toprule
        & \textbf{Precision\%} & \textbf{Recall\%} & \textbf{F-score\%}\\ \midrule
\spar{}                 & 53.1          & 36.6    & 43.3 \\
\amrdrs{}               & 46.5          & 48.2    & 47.3 \\
\textsc{Pmb\,Pipeline}  & 53.0          & 54.8    & 53.9 \\
\bottomrule
\end{tabular}
\end{table}

The results of the three baseline parsers are shown in Table~\ref{tab:baselines}. The surprisingly high score of \spar{} is explained by the fact that the first PMB release mainly contains relatively short sentences with little structural diversity. The average number of clauses per clausal form (excluding redundant REF-clauses) is 8.7, where a substantial share (approximately 3) comes from tense related clauses. 
Due to this fact, guessing temporal clauses for short sentences has a big impact on F-score.
This is illustrated by the comparison of the clausal forms in Figure\,\ref{fig:drs-compare}, where matching only temporal clauses results in an F-score of 40\%. 

\textsc{amr2drs} outperforms \textsc{spar} by a considerable margin, but is still far from optimal. This is also the case for \textsc{pmb\,pipeline}, which shows that, within the PMB, manual annotation is still required to obtain gold standard meaning representations.

\subsection{Comparing Translations}

The second purpose of \dmatch{} is checking whether translations are meaning-preserving. As a pilot study, we compare the gold standard meaning representations of German, Italian and Dutch translations in the release to their English counterparts. The results are shown in Table~\ref{tab:translations}. The high F-scores indicate that the meaning representations are often syntactically very similar, if not identical. However, there is a considerable subset of meaning representations %(10.4--13.5\%)
which are different from the English ones, indicating that there is at least a slight discrepancy in meaning for those translations. 

\begin{table}[h]
\centering
\caption{\label{tab:translations}Comparing meaning representations of English texts to those of German, Italian and Dutch translations.}
\begin{tabular}{lrrrr}
\toprule
                 & \textbf{F-score\%} & \textbf{Docs} & \textbf{F\textless 1.0} & \textbf{\% total} \\
\midrule
\textbf{German}  & 98.4             & 579              & 61                          & 10.5                 \\
\textbf{Italian} & 97.6             & 341              & 46                          & 13.5                 \\
\textbf{Dutch}   & 98.3             & 355              & 37                          & 10.4        \\
\bottomrule
\end{tabular}
\end{table}

Manual analysis of these discrepancies showed that there are several different causes for a discrepancy to arise.
In most of the cases (38\%), a human annotation error was made.
In 34\% of cases, a definite description was used in one language but not in the other. Examples are `has long hair' with the Italian translation `ha \underline{i} capelli lunghi', and `escape from prison' with the Dutch translation `vluchtte uit \underline{de} gevangenis'.
In 15\% of cases proper names were translated (e.g. `United States' and `Stati Uniti'). This is not accounted for, since we do not currently make use of grounding proper names to a unique identifier, for instance by wikification \cite{cucerzan:2007}, or by using a language-independent transliteration of names.  
In 13\% of cases the translation was either non-literal or incorrect. Examples are `Tom lacks experience' with the Dutch translation `Tom heeft geen ervaring' (lit. `Tom has no experience'), `can't use chopsticks' with the German `kann nicht mit St\"{a}bchen essen' (lit. `cannot eat with sticks'), and `remove the dishes from the table' with the Dutch translation `ruimde de tafel af' (lit. `uncluttered the table').

The mapping of clausal forms involving non-literal translations is illustrated in Figure\,\ref{fig:drs-trans}. 
This preliminary analysis shows that this comparison of meaning representations provides an an additional method for detecting mistakes in annotation. It also showed that there are cases where our semantic analysis needs to be revised and improved.

%████████████ TRANSLATIONS FIGURE ████████████
\begin{figure}[t]
\hspace*{-3mm}
\begin{tabular}{@{}c@{\kern-4mm}c@{}}
%%%%%%%%%%%%%%%%%%%%%% 14/0849
\textsmaller[1]{\sym{She~removed~the~dishes~from~the~table.}}
&
\textsmaller[1]{\sym{\kern5mm Ze~ruimde~de~tafel~af.}} 
\\[-1mm]
\scalebox{1}{
\renewcommand\arraystretch{1.1}
\drs[t]{$x_1$ ~$x_2$ ~$e_1$ ~$x_3$ ~$t_1$ \hspace{5mm}}{ 
		female$\sym{.n.02}(x_1)$\\
		remove$\sym{.v.01}(e_1)$\\
        \wsp$\sym{Time}(e_1, t_1)$\\
        \wsp$\sym{Source}(e_1, x_3)$\\
        \wsp$\sym{Theme}(e_1, x_2)$\\
        \wsp$\sym{Agent}(e_1, x_1)$\\
        $\sym{time.n.08}(t_1)$\\
        \wsp$t_1 \prec \sym{now}$\\
        dish$\sym{.n.01}(x_2)$\\
        table$\sym{.n.03}(x_3)$
}}
&
\scalebox{1}{
\renewcommand\arraystretch{1.1}
\drs[t]{$x_1$ ~$x_2$ ~$e_1$ ~$t_1$ \hspace{8mm}}{ 
		female$\sym{.n.02}(x_1)$\\
		unclutter$\sym{.v.01}(e_1)$\\
        \wsp$\sym{Time}(e_1, t_1)$\\
        \wsp$\sym{Source}(e_1, x_2)$\\
        \wsp$\sym{Agent}(e_1, x_1)$\\
        $\sym{time.n.08}(t_1)$\\
        \wsp$t_1 \prec \sym{now}$\\
        table$\sym{.n.03}(x_2)$
}}
\\[-2mm]
\textsmaller[1]{\texttt{
\begin{tabular}[t]{@{\,}l@{\,}}\toprule
\strout{\matched{\drgvar{b1} REF \drgvar{x1}}}\\
\matched{\drgvar{b1} female \posn{n}{02} \drgvar{x1}}\\
\strout{\matched{\drgvar{b5} REF \drgvar{t1}}}\\
\matched{\drgvar{b5} TPR \drgvar{t1} "now"}\\
\matched{\drgvar{b5} time \posn{n}{08} \drgvar{t1}}\\
\matched{\drgvar{k0} Agent \drgvar{e1} \drgvar{x1}}\\
\strout{\matched{\drgvar{k0} REF \drgvar{e1}}}\\
\nonmatched{\drgvar{k0} Theme \drgvar{e1} \drgvar{x2}}\\
\matched{\drgvar{k0} Time \drgvar{e1} \drgvar{t1}}\\
\nonmatched{\drgvar{k0} remove \posn{v}{01} \drgvar{e1}}\\
\strout{\nonmatched{\drgvar{b2} REF \drgvar{x2}}}\\
\nonmatched{\drgvar{b2} dish \posn{n}{01} \drgvar{x2}}\\
\matched{\drgvar{k0} Source \drgvar{e1} \drgvar{x3}}\\
\strout{\matched{\drgvar{b4} REF \drgvar{x3}}}\\
\matched{\drgvar{b4} table \posn{n}{03} \drgvar{x3}}\\
\bottomrule
\end{tabular}
}}
&
\textsmaller[1]{\texttt{
\begin{tabular}[t]{@{\,}l@{\,}}\toprule
\strout{\matched{\drgvar{b1} REF \drgvar{x1}}}\\
\matched{\drgvar{b1} female \posn{n}{02} \drgvar{x1}}\\
\strout{\matched{\drgvar{b4} REF \drgvar{t1}}}\\
\matched{\drgvar{b4} TPR \drgvar{t1} "now"}\\
\matched{\drgvar{b4} time \posn{n}{08} \drgvar{t1}}\\
\matched{\drgvar{k0} Agent \drgvar{e1} \drgvar{x1}}\\
\strout{\matched{\drgvar{k0} REF \drgvar{e1}}}\\
\matched{\drgvar{k0} Source \drgvar{e1} \drgvar{x2}}\\
\matched{\drgvar{k0} Time \drgvar{e1} \drgvar{t1}}\\
\nonmatched{\drgvar{k0} unclutter \posn{v}{01} \drgvar{e1}}\\
\strout{\matched{\drgvar{b2} REF \drgvar{x2}}}\\
\matched{\drgvar{b2} table \posn{n}{03} \drgvar{x2}}\\
\bottomrule
\end{tabular}
}}
\end{tabular}
\caption{English and Dutch non-literal translations of the document 14/0849.
Their clausal forms match each other (excl. redundant REF-clauses) with an F-score of 77.8\%. This matching is achieved by the mapping of variables \{\ntt[0]{b5}$\mapsto$\ntt[0]{b4}, \ntt[0]{b4}$\mapsto$\ntt[0]{b2}\}.}
\label{fig:drs-trans}
\end{figure}
%████████████ TRANSLATIONS END FIGURE ████████████

%\begin{verbatim}
%  42 reason: definite description used in one language but not in the other
%  41 reason: annotation error
%  19 reason: names are translated
%  14 reason: non-literal translationsl
%  12 reason: wordnet error
%   9 reason: pronouns are resolved in English but in the translation
%   3 reason: different modals in translation
%   2 reason: possessive in one language and definite description in the other language
%   2 reason: fixed already (no difference)
%\end{verbatim}

\section{Conclusions and Future Work}

%\noindent
% this is a general conclusion
Large semantically annotated corpora are rare. Within the Parallel Meaning Bank project, we are creating a large, open-domain corpus annotated with formal meaning representations. We take advantage of parallel corpora, enabling the production of meaning representations for several languages at the same time. Currently, these are languages similar to English, two Germanic languages (Dutch and German) and one Romance language (Italian). Ideally, future work would include more non-Germanic languages.

% this is about DRS, compared to AMR
The DRSs that we present are meaning representations with substantial expressive power. They deal with negation, universal quantification, modals, tense, and presupposition. 
As a consequence, semantic parsing for DRSs is a challenging task. Compared to Abstract Meaning Representations, the number of clauses and variables in a DRS is about two times larger on average. Moreover, compared to AMRs, DRSs rarely contain clauses with single variables. All non-logical symbols used in DRSs are grounded in WordNet and VerbNet (with a few extensions).
% this is about D-match
This makes evaluation using matching computationally challenging, in particular for long sentences, but our matching system \dmatch{} achieves a reasonable trade-off between speed and accuracy.

% extensions (future work)
Several extensions to the annotation scheme are possible. Currently, the DRSs for the non-English languages contain references to synsets of the English WordNet. Conceptually, there is nothing wrong with this (as synsets can be viewed as identifiers for concepts that are  language-independent), but for practical reasons it makes more sense to provide links to synsets of the original language \cite{germanet,OpenDutchWordNet,ItalWordNet,MultiWordNet}. In addition, we consider implementing semantic grounding such as wikification in the Parallel Meaning Bank.

% also about D-match: senses
%
As for other future work, we plan to include a more fine-grained matching regarding WordNet synsets, since the current evaluation of concepts is purely string-based, with only identical strings resulting in a matching clause. For many synsets, however, it is possible to refer to them with more than one word\sym{.POS.SenseNum} triple, and this should be accounted for (e.g. fox\sym{.n.02} and dodger\sym{.n.01} both refer to the same synset). 
In a similar vein, we plan to experiment with including WordNet concept similarity techniques in \dmatch{} to compute semantic distances between synsets, in case they do not fully match. 

Finally, we would like to stimulate research on semantic parsing with scoped meaning representations. Not only are we planning to extend the coverage of phenomena and the number of texts with gold-standard meaning representations for the four languages, we also aim to organize a shared task on DRS parsing for English, German, Dutch and Italian in the near future.

\section{Acknowledgements}

This work was funded by the NWO-VICI grant ``Lost in Translation -- Found in Meaning'' (288-89-003). We used a Tesla K40 GPU, which was kindly donated to us by the NVIDIA Corporation. We also want to thank the three anonymous reviewers for their comments.

% \nocite{*}
\section{Bibliographical References}
\label{main:ref}

\bibliographystyle{lrec}
\bibliography{lrec_ref}

%\section{Language Resource References}
%\label{lr:ref}
%\bibliographystylelanguageresource{lrec}
%\bibliographylanguageresource{lrec_ref}

\end{document}